\documentclass[10pt,twocolumn,letterpaper]{article}

\usepackage{wacv}
\usepackage{times}
\usepackage{epsfig}
\usepackage{graphicx}
\usepackage{amsmath}
\usepackage{amssymb}

\usepackage{xcolor}
\usepackage{comment}
\usepackage{subcaption}
\usepackage{multirow}
\usepackage{balance}
\usepackage{psfrag}

\newcommand{\todo}[1]{{\color{red}#1}}

\newcommand{\patchpair}[2]{\texttt{#1}~$|$~\texttt{#2}}

\usepackage[pagebackref=true,breaklinks=true,letterpaper=true,colorlinks,bookmarks=false]{hyperref}

\wacvfinalcopy 

\def\wacvPaperID{0022} 

\ifwacvfinal\fi
\begin{document}

\title{Aligned to the Object, not to the Image: \\
A Unified Pose-aligned Representation
for Fine-grained Recognition
}

\author{Pei Guo, Ryan Farrell \\
Computer Science Department\\
Brigham Young University\\
{ \{peiguo,farrell\}@cs.byu.edu}
}
\maketitle
\ifwacvfinal\thispagestyle{empty}\fi

\begin{abstract}

Dramatic appearance variation due to pose constitutes a great challenge in fine-grained recognition, one which recent methods using attention mechanisms or second-order statistics fail to adequately address.
Modern CNNs typically lack an explicit understanding of object pose and are instead confused by entangled pose and appearance.
In this paper, we propose a unified object representation built from a hierarchy of pose-aligned regions.
Rather than representing an object by regions aligned to image axes, the proposed representation characterizes appearance relative to the object's pose using
pose-aligned patches whose features are robust to variations in pose, scale and rotation. 
We propose an algorithm that performs pose estimation and forms the unified object representation as the concatenation of hierarchical pose-aligned regions features, which is then fed into a classification network.
The proposed algorithm surpasses the performance of other approaches, increasing the state-of-the-art by nearly 2\% on the widely-used CUB-200~\cite{WahBWPB_Tech2011} dataset and by more than 8\% on the much larger NABirds~\cite{VanHornBFHBIPB_CVPR2015} dataset.
The effectiveness of this paradigm relative to competing methods suggests the critical importance of disentangling pose and appearance for continued progress in fine-grained recognition.

\end{abstract}

%
%

\section{Introduction}
\label{sec:intro}

\begin{figure}       \includegraphics[width=1\linewidth]{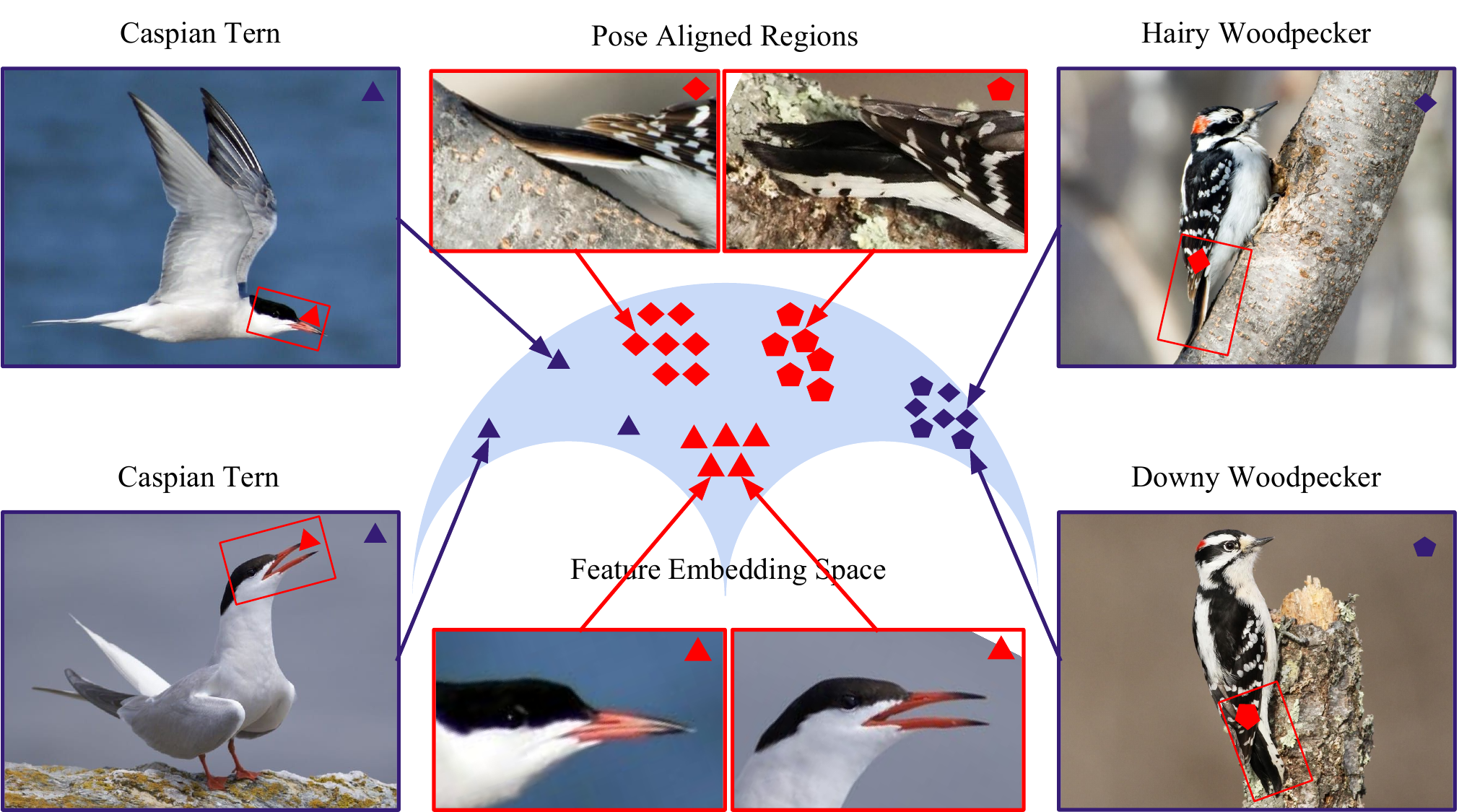}
       \caption{\textbf{Motivation for Pose-Aligned Regions.}  Two terns of the same species, but in different poses, have dramatically different appearances while two different species of woodpecker appear nearly identical except for the barring pattern on the outer tail (and the shape of the beak). The large intra-class variance and small inter-class variance make the feature space distance inaccurately reflect the true class relationships. Such observations motivate the use of \emph{pose-aligned regions} that disentangle intrinsic part appearance from variations in object pose, leading to a feature space that facilitates correctly classifying the species or fine-grained category.}
\label{fig:pairs} 
\end{figure}

What makes fine-grained visual categorization (FGVC), commonly referred to as fine-grained recognition, different from general visual categorization?  
One important distinction lies in the difficulty of the datasets. 
General-purpose visual categorization often involves the classification of everyday objects, such as chairs, bicycles and dogs, which are easy for humans to identify. Fine-grained recognition, on the other hand, consists of more detailed classifications such as identifying the species of a bird.  This is extremely difficult for non-expert humans as it requires familiarity with domain knowledge and hundreds of hours of training.
Computer algorithms for fine-grained recognition have the potential to be far more accurate than most humans and can thus benefit millions of people by providing services like species recognition through mobile applications~\cite{leafsnap_eccv2012,berg2014birdsnap,Merlin_Bird_ID}.

An intrinsic and readily observed quality of fine-grained recognition is small inter-category variance coupled with large intra-category variance.
 Discriminative features of two visually similar categories often lie in a few key locations, while the appearances of the objects from the same category be dramatically different due simply to pose variation.
The entanging of appearance and pose presents a great challenge and motivates the need for stable appearance features, ones that are nearly invariant to variations in pose, scale and rotation.

\begin{figure}       
\centering
\includegraphics[width=0.9\linewidth]{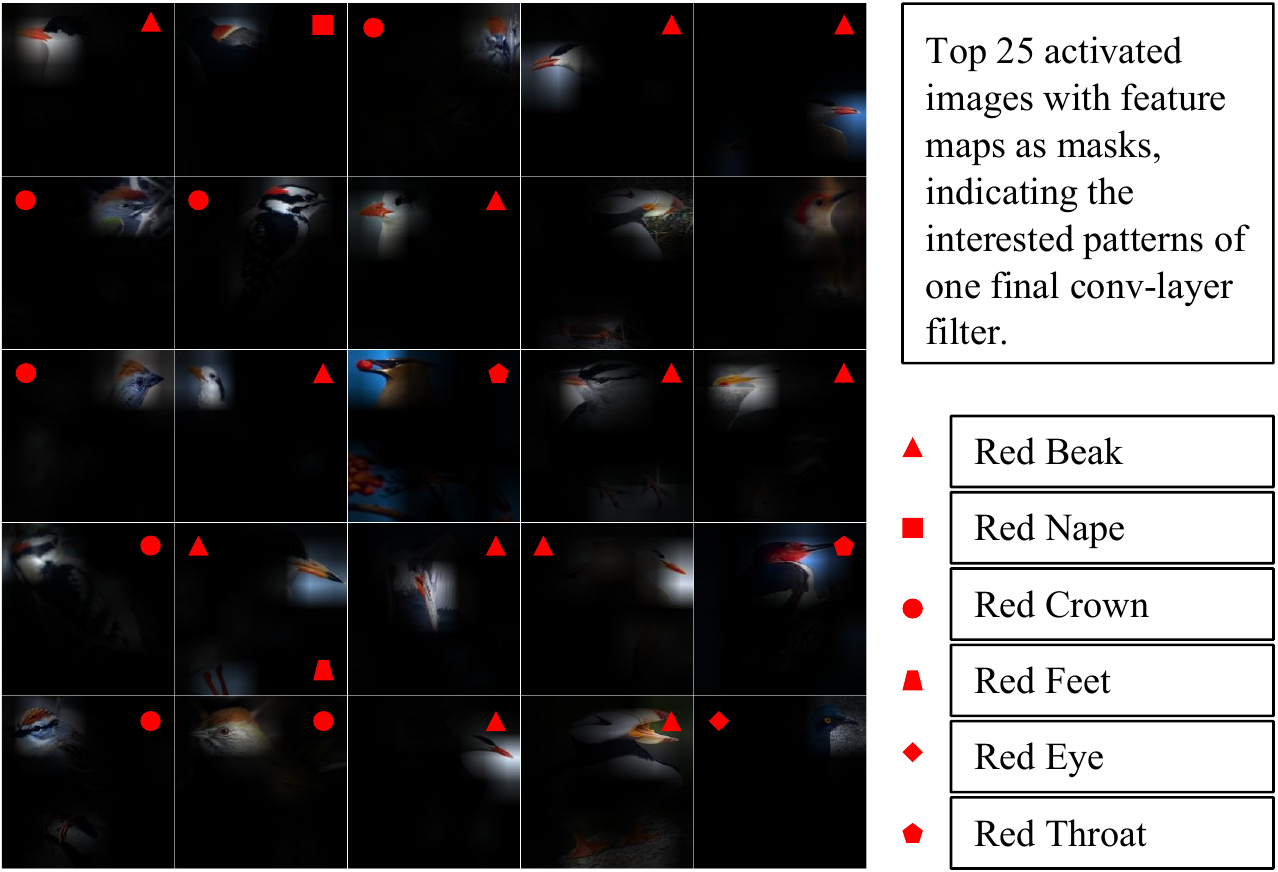}
       \caption{\textbf{Filter Visualization as Proof that CNNs Lack Part-Awareness.} We visualize the 25 top-activated images for a filter from the ResNet-50 model's final convolutional layer. Modern CNNs are purely appearance-based and lack part-awareness -- as a result, different semantic parts (beak, feet, crown, throat, nape or eyes) that are red in color can all activate this filter, causing confusion for the classifier when such semantic parts would be discriminative. Pose-aligned regions can eliminate this problem by disentangling parts and appearance.}
\label{fig:visualization} 

\end{figure}

It's almost instinct for humans to identify and visually compare key locations across objects in different poses, establishing correspondences.
Convolutional neural networks, however, struggle on this task because the convolutional mechanisms are purely appearance-based and lacks an understanding of the pose or geometry.
The built in pooling mechanisms can tolerate a certain amount of scale and rotation variation~\cite{cohen2016steerable,cohen2016group,marcos2016rotation,weiler2018learning,cohen2018spherical}, but exactly how much is still largely an open question~\cite{ruderman2018learned}.
We show this in Figure~\ref{fig:visualization} via the visualization of some final convolutional layer responses.
We show the top-activated images together with the feature map as a masks.
It's evident that this convolutional filter is attuned to red beaks.
However, due to its lack of part-awareness, this filter also fires strongly at visually similar parts such as red crowns, red throats, red eyes, \etc.
This causes confusion for the classifier because of the noisy entangled part-appearance representation.

In the feature embedding space, dramatic pose variation would make images of the same category farther separated and images of visually-similar categories appear closer together as shown in Figure~\ref{fig:pairs}.
It is therefore vital that pose-aligned regions, which explicitly factor out pose variation, should be the building block of the disentangled image representation.

Recent efforts in fine-grained recognition have largely focused on two directions.
One is second-order statistics based algorithms\cite{LinRM_ICCV2015,GaoBZD_CVPR2016,KongF_CVPR2017,CuiZWLLB_CVPR2017}.
Representative works include Bilinear Pooling\cite{LinRM_ICCV2015} and its reduced-memory variants~\cite{GaoBZD_CVPR2016,KongF_CVPR2017} or thosethat extends to higher-order statistics~\cite{CuiZWLLB_CVPR2017}.
The idea is to project the features onto a higher-order space where they can be linearly separated.
Second-order statistics methods have both sound theoretical support and work well in practice.
However, they look at the image globally, and thus having little hope of finding subtle highly-localized differences.
Also, they lack interpretability and insights for further improvement.

The other direction is attention-based methods~\cite{FuZM_CVPR2017,LamMT_CVPR2017,LiuWWDL_AAAI2017,SermanetFR_ICLR2015,XiaoXYZPZ_CVPR2015,ZhaoWFPY_TMM2017} that use subnetworks to propose possible discriminative regions to attend to.
However, the regions proposed by these networks are often weakly-supervised by some heuristic loss function, lacking proof that they really attend to the right position.  
Both of these directions suffer from a lack of pose awareness and moreover the entanglement of pose and appearance features limits their performance.
Moreover, training data is often scarce in the long-tailed distributions seen in many fine-grained domains; in such cases, both techniques suffer as the limited training imagery
does not adequately span the space of pose and viewing angle for each category, hindering their ability to recognize any species in any pose.

Based on the above observations, we propose to disentagle pose and appearance via a unified object representation built upon pose-aligned regions, defined as rectangular patches defined relative to two keypoints anchors.
The final object representation is an aggregation of the features across all of the pose-aligned regions.
This representation comprises a pose-invariant and over-complete basis of features from multiple scales.
We contrast the pose-aligned regions with weakly-supervised regions that are generated in a purely data-driven fashion and with ``axis-aligned'' rectangular bounding boxes centered around a keypoint or landmark.
The features from these types of regions are subject to the variation of pose, scale and rotation.
We experimentally demonstrate that axis-aligned regions are inferior pose-aligned regions  with respect to classification accuracy (see Figure~\ref{fig:scatter}).

\begin{figure*}[t]
        \centering
        \includegraphics[width=1.0\linewidth]{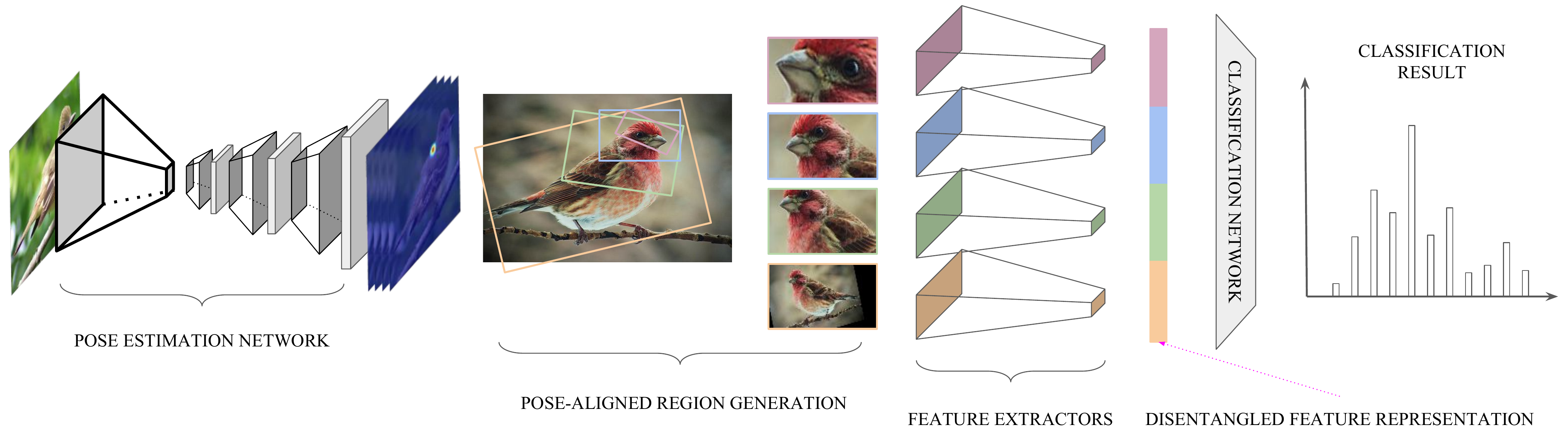}
        \caption{ {\textbf{Overview of the Proposed Framework for Fine-grained Recognition.}}  We first apply a pose-estimation network to the image for keypoint detection. Pose-aligned regions are then extracted from the image using the predicted keypoint locations.  We then extract features from the individual regions using region-specific networks.  The concatenated features collectively form a unified multi-scale representation that is invariant to pose, scale and rotation change. This representation is then fed into a classification network for the final fine-grained classification. 
        }
        \label{fig:big-picture}
\end{figure*}

To automate the process of applying the unified object representation to fine-grained recognition, we propose an algorithm that first does pose-estimation for keypoint detection, enabling the generation of pose-aligned region features.
The local features from these aligned regions, regions of varying size/scale relative to the object, are concatenated to comprise the unified representation for the input image and are then fed into a classification network to produce a final classification prediction.
We call the proposed algorithm PAIRS: Pose and Appearance Integration for Recognizing Subcategories.
It achieves state-of-the-art results on two key fine-grained datasets: CUB-200-2011~\cite{WahBWPB_Tech2011} and NABirds~\cite{VanHornBFHBIPB_CVPR2015}.
Keypoint annotations are used only during training time.
Considering the annotation cost, keypoint annotations are actually less expensive and time-consuming than collecting additional data samples because keypoints can be annotated by human non-experts wheras fine-grained image category annotation requires the consensus of multiple domain experts.



%
%

\section{Background and Related Work}

Fine-grained visual categorization (FGVC) lies between generic category-level object recognition like the VOC~\cite{EveringhamMLIJA_IJCV2015}, ImageNet~\cite{RussakovskyDSKSMHKKBBF_IJCV2015}, COCO~\cite{LinMBHPRDZ_ECCV2014}, \etc~and instance-level classification like facial recognition and other visual biometrics. The challenges of FGVC are many-fold: differences between similar species are often subtle and highly-localized and thus difficult even for (non-expert) humans to identify. Dramatic pose changes introduce great intra-class variance. Generalization also becomes an issue as the network struggles to find truly useful and discriminative features. 

FGVC has drawn wide attention in the computer vision community. Some early works include~\cite{DuanPCG_CVPR2012,FarrellOMZDD_ICCV2011,LiuKJB_ECCV2012,YaoBF_CVPR2012,YaoKF_CVPR2011,ZhangFD_CVPR2012,ZhangFID_ICCV2013}. Birdlet, a volumetric poselet representation is proposed to account for the pose and appearance variation in~\cite{FarrellOMZDD_ICCV2011}. \cite{ZhangFID_ICCV2013} further proposes two pose-normalized descriptors based on computationally-efficient 
deformable part models. Although these early works seek to integrate pose and appearance like our method does, they rely heavily on hand-engineered descriptors thus have limited success on classification accuracy. 


Our work is related to part-based CNN models~\cite{BransonVPB_BMVC2014,HuangXTZ_CVPR2016,KrauseGDLF_ICPR2014,ZhangDGD_ECCV2014,ZhangSGD_ICLRWorkshop2016,DiL_CVPR2015}, which seek to decompose the object into semantic parts. \cite{ZhangDGD_ECCV2014} first employs an object detection framework -- R-CNN~\cite{Girshick_CVPR2015} for object and part detection.
Part-Stacked CNN~\cite{ShaoliH_CVPR2016} proposes a fully convolutional network for keypoint detection and a two-stream convolutional network for object and part level feature extraction. Deep LAC~\cite{DiL_CVPR2015} proposes a valve linkage function for back-propagation chaining and form a deep localization, alignment and classification system. \cite{ZhangSGD_ICLRWorkshop2016} introduces an end-to-end learning framework for joint learning of pose estimation, normalization and recognition. These models are all based on a limited number of single keypoint patches, which could be poorly-aligned in the presence of pose and viewpoint variance. 

Perhaps our work is mostly related to POOF~\cite{BergB_CVPR2013}, which also uses keypoint pair patch. Our algorithm is different from theirs as we automatically detect keypoints instead use
ground truth ones. Also the POOF approach computed 5000 patches with corresponding features in order to produce the final classification, we're computing 35-70.

There are also works targeting the object alignment problem. Unlike previous methods which rely on detectors for part localization, \cite{GavvesFSST_ICCV2013,GavvesFSST_IJCV2015} proposes to localize distinctive details by roughly aligning the objects using just the overall shape. 
Spatial transformer network~\cite{JaderbergSZK_NIPS2015} introduces a differentiable affine transformation learning layer to transform and align the object or part of interest. 

Another direction in fine grained recognition is feature correlation and kernel mapping.
\cite{LinRM_ICCV2015}~proposes a bilinear pooling layer to compute a second order polynomial kernel mapping on CNN features. Many works has followed this simple paradigm~\cite{GaoBZD_CVPR2016,KongF_CVPR2017,CuiZWLLB_CVPR2017}.
Compact bilinear pooling~\cite{GaoBZD_CVPR2016} proposes a  compact representation to approximate the polynomial kernel, reducing memory usage.
Low-rank bilinear pooling~\cite{KongF_CVPR2017} represents the covariance features as a matrix and applies a low-rank bilinear classifier.
Kernel pooling~\cite{CuiZWLLB_CVPR2017} proposes a general pooling framework that captures higher order interactions of features in the form of kernels. This line of works achieves relatively good results with weakly supervision, however, they attend to the whole image globally, lacking part-level information discovery mechanism. This limites their success in further accuracy improvement.

Inspired by human attention mechanism, 
many attempts have been made to guide the attention of the CNN model to informative object parts. 
Works along this direction include~\cite{FuZM_CVPR2017,LamMT_CVPR2017,LiuWWDL_AAAI2017,SermanetFR_ICLR2015,XiaoXYZPZ_CVPR2015,ZhaoWFPY_TMM2017}.
\cite{HeliangZ_ICCV2017} proposes a multi-attention convolutional
neural network (MA-CNN), where part generation and
feature learning can reinforce each other.
\cite{LamMT_CVPR2017} leverages long short term memory networks (LSTM) to unify new patch candidates generation and informative part evaluation. This work establishes the current state-of-the-art performance on CUB-200-2011~\cite{WahBWPB_Tech2011} dataset, achieving an accuracy of 87.5\%  with part annotations. The key difference is our PAIRS representation integrates pose and appearance information and achieves multi-level attention over semantic object parts explicitly at the same time.


\section{\underline{\bf PAIRS} - Pose and Appearance Integration}

\begin{figure}[h]
        \centering
       \includegraphics[width=1\linewidth]{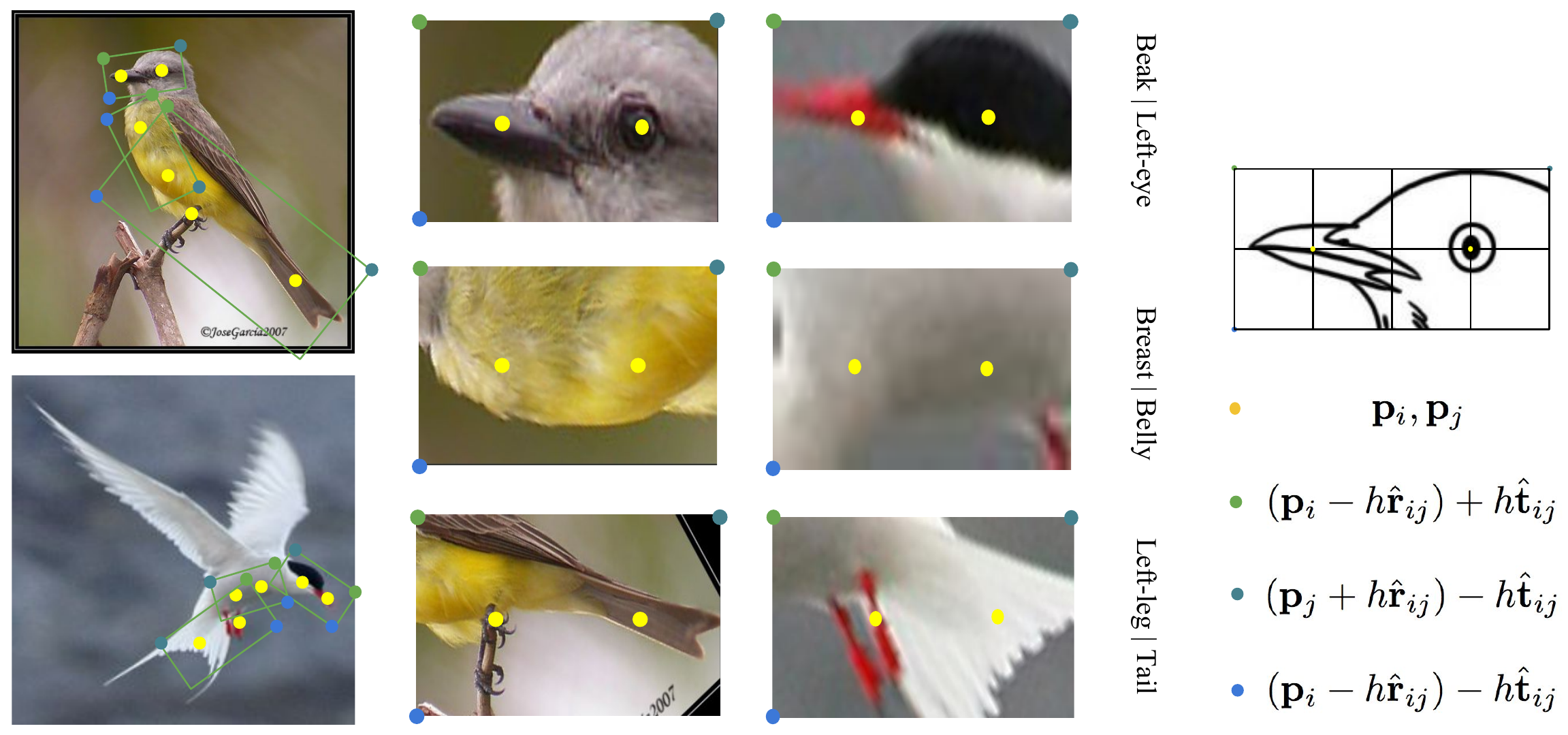}
        \caption{ \textbf{Pose-aligned Patch Generation.} We show the pose-aligned patch generation in this graph. For each pair of keypoints, we fit a rectangle bounding box whose corners are calculated as in (b). Objects of different poses and from different viewpoint can be compared directly by the proposed keypoint pair patch as shown in (a). Details are described in Section~\ref{sec:patch_gen}.}
\label{fig:pose_align} 
\end{figure}

We illustrate our algorithm pipeline in Figure~\ref{fig:big-picture}. 
Firstly, we propose a simple yet effective fully convolutional neural network for pose estimation. 
We follow the prevailing modular design paradigm by stacking convolutional blocks that have similar topology. 
we show our pose estimation network achieves supreme results on the CUB-200 dataset both qualitatively and quantitatively.
Secondly, 
given detected keypoint locations, 
a rectangle bounding box enclosing each keypoint pair is cropped from the original image and similarity-transformed to a uniform-sized patch (Figure~\ref{fig:pose_align}),
such that both keypoints are at fixed position across different images.
As the representation is normalized to the keypoint locations, the patches are well-aligned, independent of the pose or the viewer's angle.
Thirdly, 
we train separate CNN models as feature extractors for the pose-aligned patch representation.
Lastly,
we explore different classification architectures for the unified representation based on the assumption that part contribution should vary for different images and classes. 
We find surprisingly that the Multi-Layer Perception (MLP), while being the most simple method, achieves the best final classification accuracy.



\subsection{Pose Estimation Network}

Pose estimation networks usually follow one of two paradigms for prediction.
The first is to directly regress discrete keypoint coordinates, e.g. $(x_i, y_i)$.  Representative approaches include~\cite{ToshevC_CVPR2014}.
The alternate approach~\cite{TompsonJLB_NIPS2014} instead uses a two-dimensional probability distribution heat map to represent the keypoint location. We call this resulting multi-channel probability distribution matrix \emph{pose tensor}.

In this paper, we adopt the second strategy, proposing a fully convolutional network to produce the structured output distribution.
Specifically, we take a pretrained classification network and remove the final classifier layer(s), retaining what can be seen as an encoder network that encodes strong visual semantics. We follow the prevailing modular design to stack repeated building blocks to the end of the network. The resulting block consists of one upsampling layer, one convolutional layer, one batch normalization layer and one ReLU non-linearity layer. The parameter-free bilinear interpolation layer is used for upsampling. The convolutional layer has 1x1 kernel and reduces the input channel size to half. Additionally, a final convolutional layer and upsampling layer are added to produce the pose tensor. There are many modifications one can make to enhance this basic model, including using larger 3x3 kernels, adding more convolutional layers to the building block,  adding residue connection to each block, stacking more building blocks, and using a learnable transposed convolutional layer for upsampling. We find these structures provide only limited improvement but introduce more parameters, so we prefer this simpler architecture.

\subsection{Patch Generation} \label{sec:patch_gen} 

Historically, part-based representations would model parts either as rectangular regions \cite{ZhangDGD_ECCV2014,FelzenszwalbGMR_PAMI2010} or keypoints.  
Keypoints are convenient for pose-estimation. 
However, the square or rectangular patches, each centered on a given keypoint and extracted to characterize the part's appearance, are far from optimal in the presence of rotation or general pose variation.
We instead, propose to use keypoint pairs as anchor points in extracting pose-aligned patches.

Given two keypoints $\vec{p}_i = (x_i, y_i)$ and $\vec{p}_j = (x_j, y_j)$, we define the vectors $\vec{r}_{ij}=p_j - p_i$, and $\hat{\vec{r}}_{ij}=\vec{r}_{ij}/||\vec{r}_{ij}||_2$.  We also define the vector $\hat{\vec{t}}_{ij} = \hat{\vec{z}} \times \hat{\vec{r}}_{ij}$, a unit vector perpendicular to $\vec{r}_{ij}$, and the distances $d=||\vec{r}_{ij}||_2$ and $h=d/2$ for convenience.
We seek to extract a region around $\vec{p}_i$ and $\vec{p}_j$ that is aligned with $\vec{r}_{ij}$ and has dimensions $2d \times d$.  The four corners of this rectangular region are then given by:
\begin{equation}
\begin{bmatrix}
~(\vec{p}_i-h \hat{\vec{r}}_{ij}) + h \hat{\vec{t}}_{ij}~ &
~(\vec{p}_j+h \hat{\vec{r}}_{ij}) - h \hat{\vec{t}}_{ij}~ \\
~(\vec{p}_i-h \hat{\vec{r}}_{ij}) - h \hat{\vec{t}}_{ij}~ &
~(\vec{p}_j+h \hat{\vec{r}}_{ij}) + h \hat{\vec{t}}_{ij}~
\end{bmatrix}
\end{equation}

A similarity-transform is computed to extract the pose-normalized patch.
Patches generated in this way contain stable pose-aligned features -- features near these keypoints appear at the same location in the given patch different images independent of the object's pose or the camera viewing angle.

\subsection{Patch Feature Extraction}

 A separate patch classification network is trained for each posed-aligned \patchpair{A}{B} patch as feature extractor. 
The softmax output from each network are concatenated as the representation for the input image. Alternatively, the final convolutional layer output after pooling can also be used and the result is comparable. We find that symmetric parts can help reduce the overall classifier number by nearly 50\%, which is described in Section~\ref{sec:patch_network}.
The proposed patch representation can be seen as spatial pyramid that explicitly captures information of different parts at multiple spatial scales on the object.


\subsection{Classification Network}

To fully utilize the abundant patch representations, we explore different ways to form a strong classification network. 
Based on the assumption that only a small fraction of the patches contains discriminative information and patches contribution should be weighted, we explores the following strategies.

1). \textbf{Fixed patch selection}: take the average score for a fixed number of top ranking patches. This strategy can also predicts the potential of our PAIRS representation.

2). \textbf{Dynamic patch selection}: employ the sparsely gated network~\cite{ShazeerMMDLHD_ArXiv2017} to dynamically learn a selection function to select a fixed number of patches for each input. 

3). \textbf{Sequential patch weighting}: apply a Long Short Term Memory Networks (LSTM) to reweigh different patch features in a sequential way.

4). \textbf{Static patch weighting}: learn a Multi-Layer Perceptron network, which essentially applies a non-linear weighting function to aggregate information from different patches. 

We find surprisingly that the MLP network, while being the simplest network architecture, achieves the best accuracy out of all the attempts we made. Details are included in the experiment section.

\section{Experimental Evaluation}

\begin{table*}[h]
\centering
\caption{\textbf{PCK comparison} }
\label{pck-table}
\begin{tabular}{l | c | c | c | c | c | c | c | c }
\hline
& back & beak & belly & breast & crown & forehead & left-eye & left-leg\\
\hline
\hline
\cite{HuangXTZ_CVPR2016} & 80.7 & 89.4 & 79.4 & 79.9 & 89.4 & 88.5 & 85.0 & 75.0 \\
\hline
\cite{ZhangSGD_ICLRWorkshop2016} & 85.6 & 94.9 &81.9 & 84.5 & 94.8 & 96.0 & 95.7 & 64.6 \\
\hline
Ours  & \bf{91.3} & \bf{96.8} & \bf{89.0} & \bf{91.5} & \bf{96.9} & \bf{97.6} & \bf{96.9} & \bf{80.2}\\
\hline
& left-wing & nape & right-eye & right-leg & right-wing & tail & throat & \textbf{Overall} \\
\hline
\cite{HuangXTZ_CVPR2016} & 67.0 & 85.7 & 86.1 & 77.5 & 67.8 & 76.0 & 90.8 & 86.6\\
\hline
\cite{ZhangSGD_ICLRWorkshop2016} &  67.8 & 90.7 & 93.8 & 64.9 & 69.3 & 74.7 & 94.5 & N/A\\
\hline
Ours & \bf{76.8} & \bf{94.6} & \bf{97.4} & \bf{80.3} & \bf{75.3} & \bf{83.6} & \bf{97.4} & \bf{90.5}\\
\hline
\hline
\end{tabular}
\end{table*}

We test our algorithm on two datasets, the CUB-200-2011 dataset and the NABirds dataset.
The CUB-200-2011 contains 200 species of birds with 5994 training images and 5794 testing images. 
The NABirds dataset has 555 common species of birds in North America with a total number of 48,562 images.
Class labels and keypoint locations are provided in both datasets.


\subsection{Keypoint Prediction Performance}

We use PCK (Percentage of Correct Keypoints) score to measure the accuracy of keypoint prediction results.
A predicted keypoint ($p$) is ``correct'' if its within a small neighborhood of the ground truth location ($g$), or equally speaking,  
$$||p - g||_2 \le c * \max(h, w)$$ 
$c$ is a constant factor and $\max(h,w)$ is the longer side of the bounding box.

\begin{figure}[h]
        \centering
\includegraphics[width=0.8\linewidth]{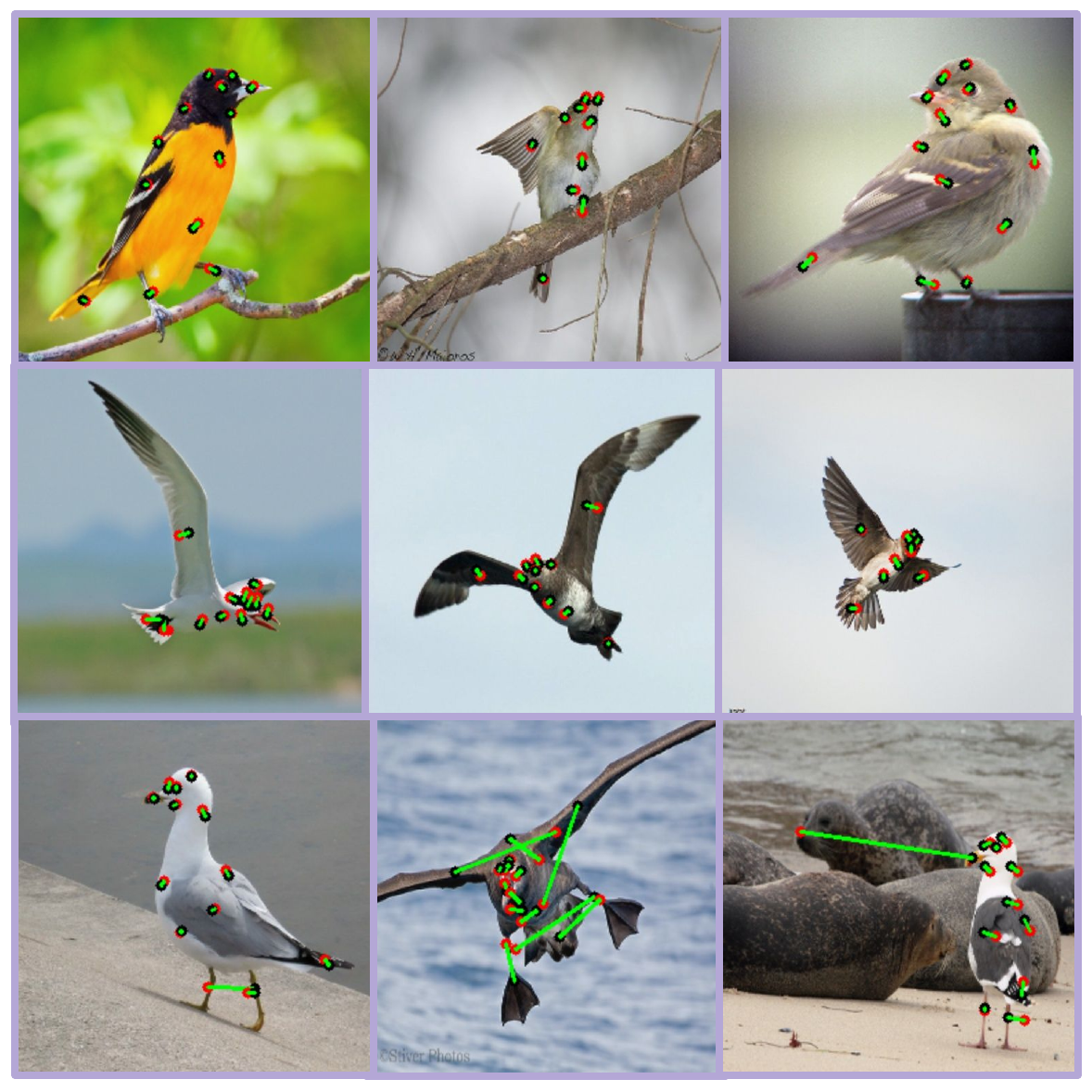}
        \label{fig:kp_examples}
        \caption{ {\textbf{Keypoint Localization Results}}
        The top two rows show the keypoint detection results of our pose estimation network. Red dots represent the predicted location and black dots are the ground truth locations. Three failure patterns are shown in the third row. The first one confuses left and right because of the visual similarity between symmetric parts. Dramatic and rare pose causes degradation for keypoint localization as seen in the middle one. The last one shows an interesting case where the nose of a seal is mistakenly predicted as the gull's beak.
         }
\end{figure}

We evaluate our pose estimation network on CUB-200 and compare our PCK score with the others in Table ~\ref{pck-table}. 
We achieve highest score on all 15 keypoints with considerable leading margin.
We do especially well on legs and wings where other models struggle to make precise prediction.
Some visualization results are shown in~\ref{fig:kp_examples}

Although we localize wings and legs better than baselines, they still have the lowest PCK in our model.
This is caused by dramatic pose change as well as the appearance similarity between symmetric parts.
We note that using keypoints to denote the wings isn't always appropriate.
Because wings are two dimensional planar parts that spread over a relatively large area.
Designating a keypoint to the wing can be obscure,
because it is not easy to decide which point represents the wing location better. 
In fact, the ground truth keypoint location of the CUB dataset is the average of five annotators' results and 
it's even hard for them to reach a consensus.

\subsection{Patch Classification Network} \label{sec:patch_network}
We adopt the ResNet-50 architecture as the patch classification network due to its high performance and compact GPU footprint.
Alternate architectures like VGG and Inception can easily be adapted. 
We now discuss two considerations which facilitate training the patch classification network.

\noindent\textbf{Symmetry.} 
For a given object with $n$ keypoints, the total number of patches to be classified is:
$$ \begin{pmatrix} n \\ 2  \end{pmatrix} = \frac{n(n-1)}{2} = \mathcal{O}(n^2)$$
which increases quadratically with $n$.
Most real world objects show some kind of symmetry. Due to the visual
similarity inherent in symmetric pairs keypoints (for example, right and left eyes, wings and feet), we treat
each pair as a hybrid keypoint in the patch generation process.  Many real-world objects, however, like birds, cats, cars, \etc~are symmetric in appearance.  Based on this observation, we propose to merge the patches for a symmetric pair of keypoints into a hybrid patch, \eg~\patchpair{left-eye}{tail} and \patchpair{right-eye}{tail} can be merged into the hybrid \patchpair{eye}{tail} pair.

As a result, the total number of patch classification networks is reduced from 105 to 69 for the CUB dataset; on the NABirds dataset, the number is reduced from 55 to 37.

\noindent\textbf{Visibility.} 
Due to self-occlusion or foreground-occlusion, not all keypoints are visible in the image. Previous works~\cite{HuangXTZ_CVPR2016} would eliminate patches with invisible keypoints to purify the input data. 
Contrarily, we find that this would hurt the performance of the patch classifiers. Details for comparison can be found in Figure~\ref{fig:many_graph}.
We believe this degradation is caused by the shrinkage of effective training set size. 
This is a similar finding with~\cite{KrauseSHZTDPF_ECCV2016} that noisy but abundant data consistently outperforms clean but limited-sized data. 
Additionally, the pose estimation network would make a reasonable guess even if the keypoint is invisible. 
So during patch classifier training, all keypoints are considered visible by taking the maximally activated location.

\subsection{Classification Network}

Based on the assumption that image patches should contribute differently to classification. Four different strategies are explored and we describe details of them in this section.

\begin{figure*}[h]
        \centering
       \includegraphics[width=0.45\linewidth]{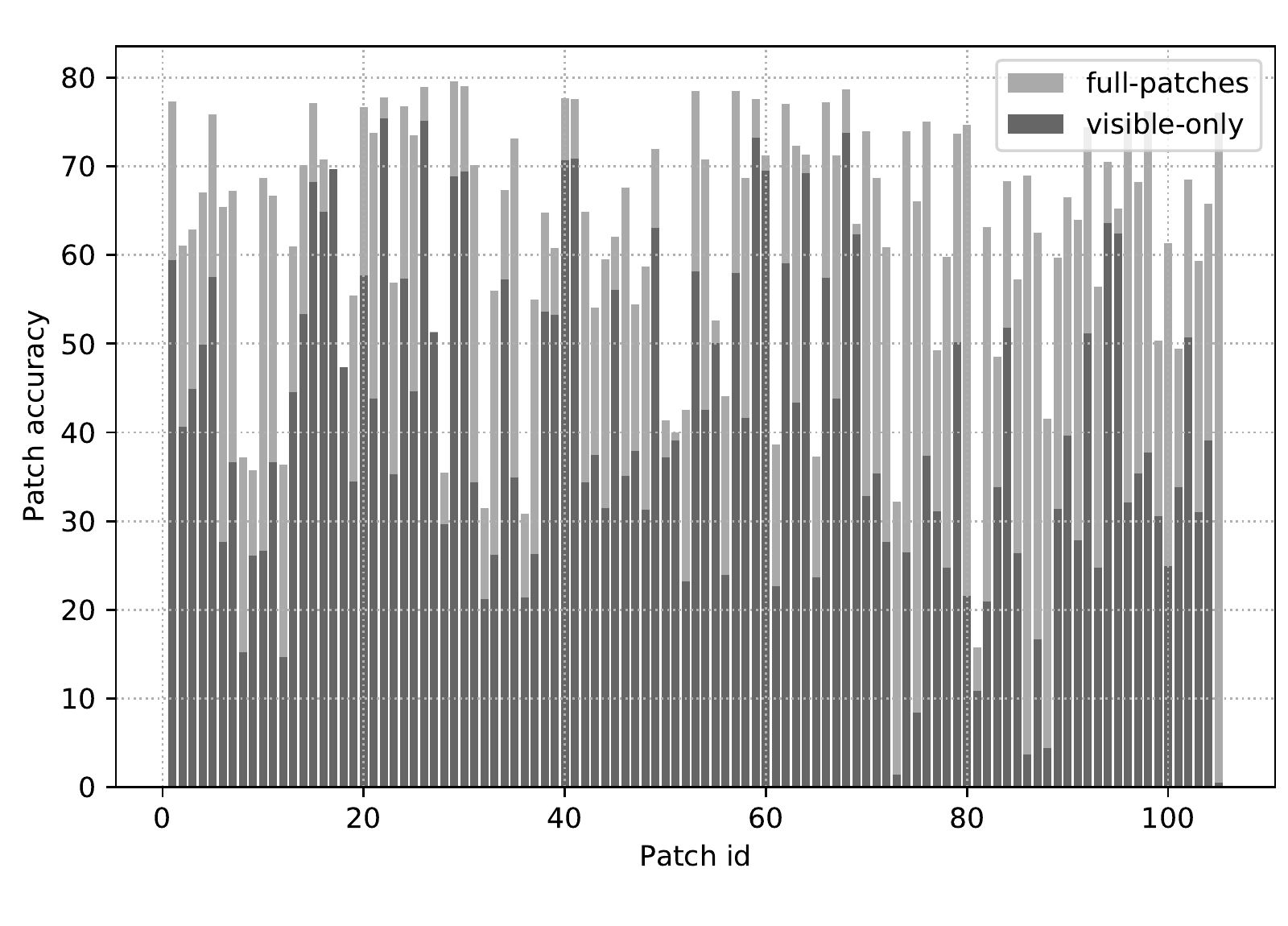}
        \includegraphics[width=0.45\linewidth]{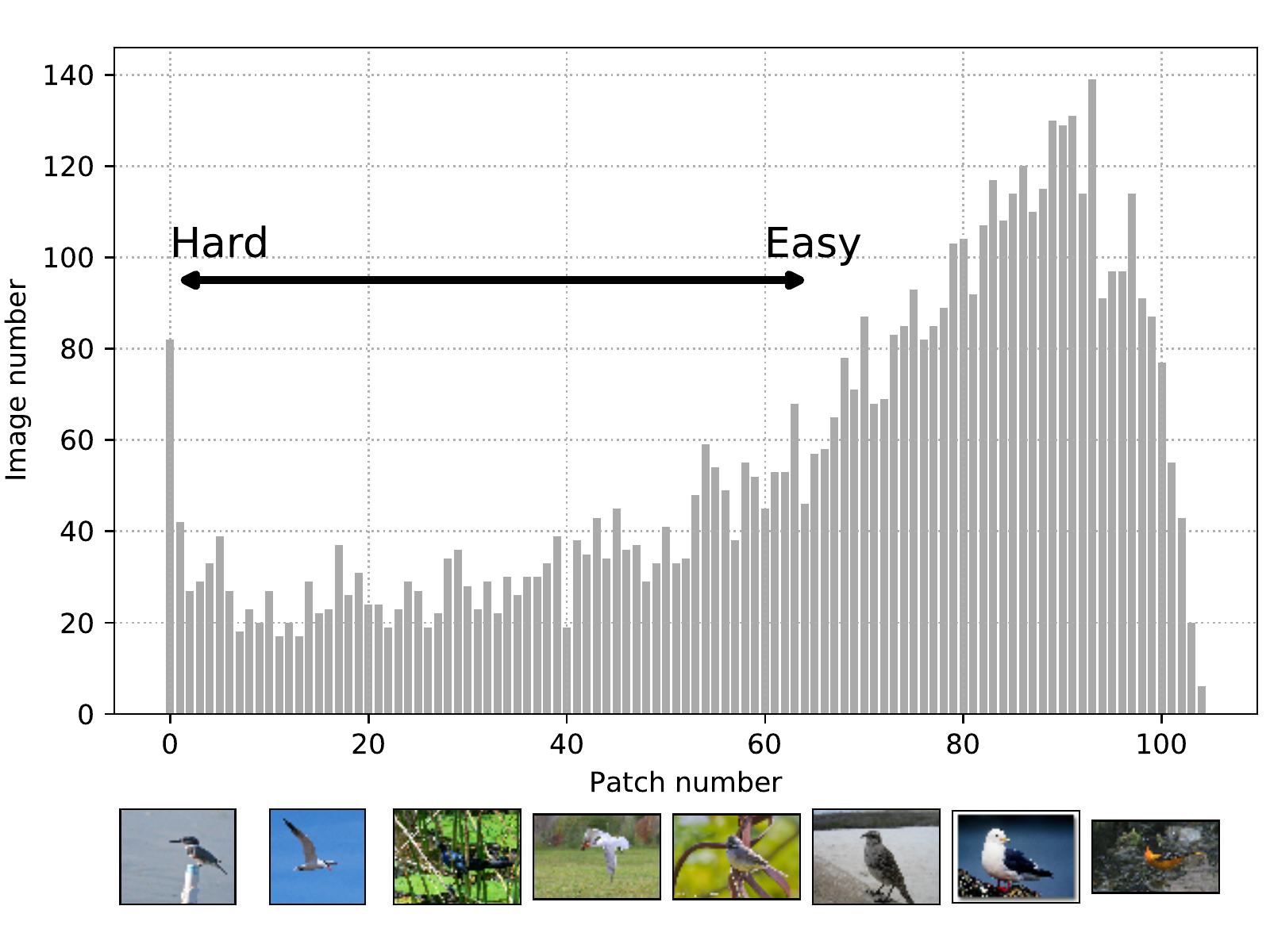}
        \includegraphics[width=0.45\linewidth]{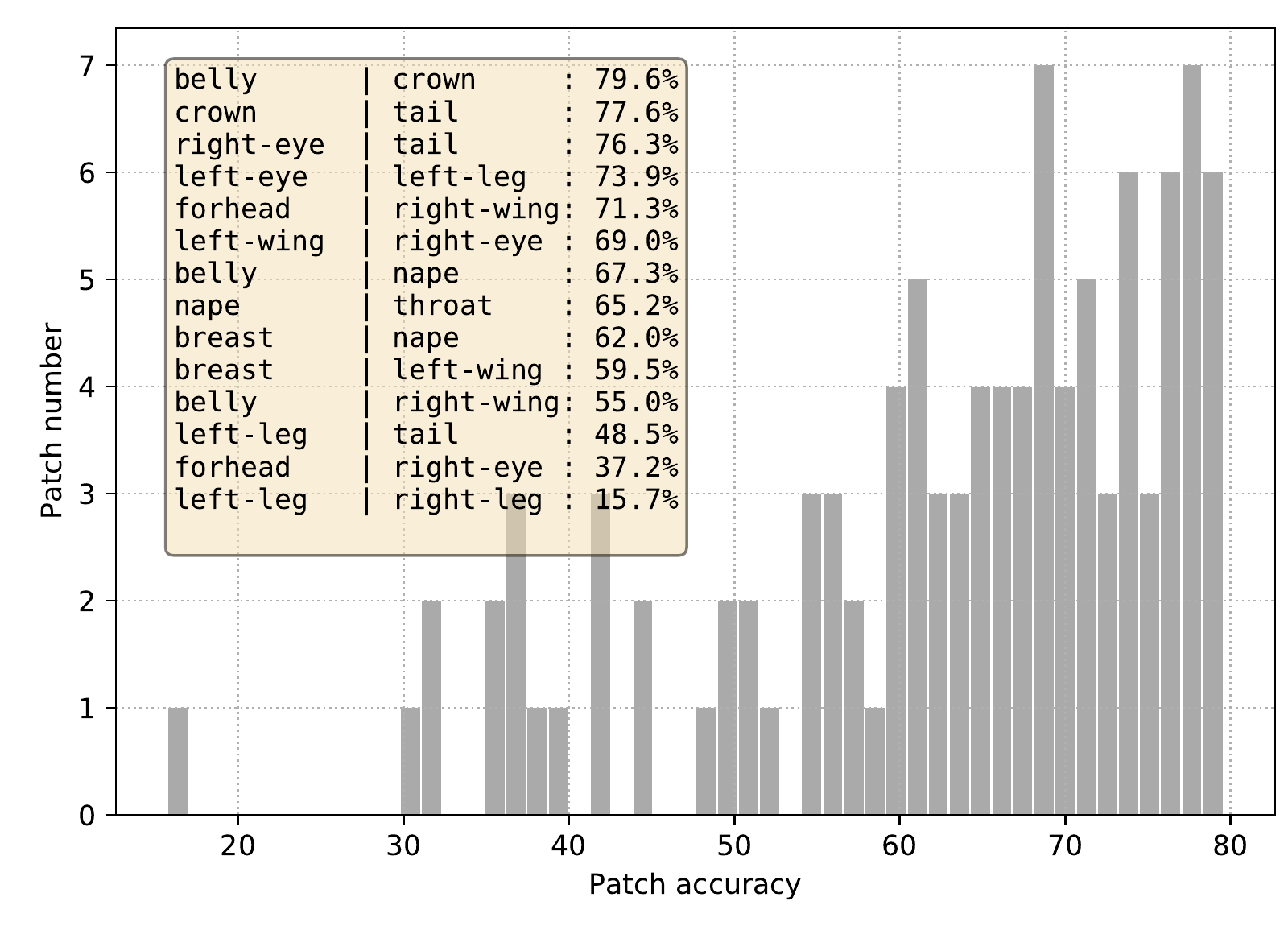}
        \includegraphics[width=0.45\linewidth]{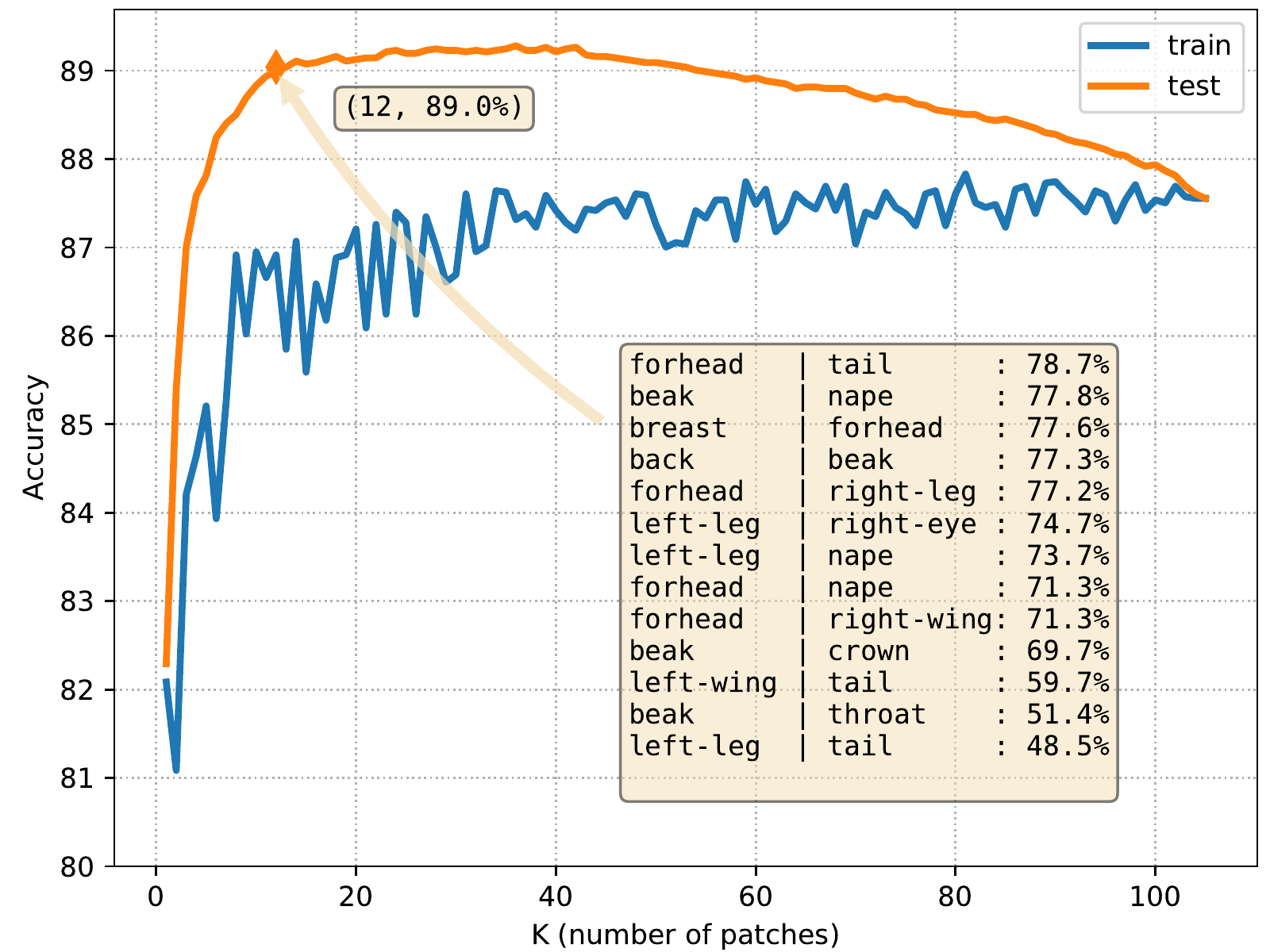}
       
        \caption{\textbf{Result visualization.} Top left: we show patch classification network accuracies using two strategies, visible keypoint patch and full keypoint patch separately. This verifies that treating all keypoints as visible will help improve the patch classifier's accuracy. Top right: hard case mining by correct prediction patch number. In the bottom we show sample images from hard to easy. Bottom left: distributions of patch classifier performance. Some examples are shown in the text box. Bottom right: Beam search results using two strategies, patch finding on training set (blue) and testing set (orange). The later one is purely for the estimation of the potential of the pair representation.
}
\label{fig:many_graph} 
\end{figure*}

\noindent\textbf{Fixed patch selection.}
We assume only few patches contains useful information and others may merely act as noise.
We propose a fixed patch selection strategy to keep the best $k$ patches.A greedy search algorithm would evaluate each $n$ choose $k$  combinations for $k$  from 1 to $n$. 
The number of evaluations needed for this algorithm is 
The complexity grows in the order of $n!$ and quickly becomes intractable. 
We thus employ the \emph{beam search} algorithm.
Instead of greedily searching the whole parameter space, we only keep a fixed $k$ combinations each iteration and build our search path based only on previously learned $k$ patch combinations. 
Out of curiosity,
we also do beam search on the testing set alone. 
This operation, although invalid, provides some insights in the potential of our pose aligned patch representation. 
The results are shown in Figure~\ref{fig:many_graph}.
Our observations is that without overfitting, the potential of fixed patch selection should be well above 89\%, compared to the current state-of-the-art~\cite{LamMT_CVPR2017} 87.5\%. Notably, a simple average over all strategy can achieve 87.6\%.

\noindent\textbf{Dynamic patch selection.} As an alternative attempt 
we experiment with is the sparsely gated network~\cite{ShazeerMMDLHD_ArXiv2017} for dynamic patch selection. 
Different from the beam search algorithm which selects fixed patches for each input,
the gated network would select different combinations depending on the input.
A tiny network is trained to predict weights for each patch and an explicit sparsity constraint is exposed on the weight to only allow $k$ non-zero elements. 
A Sigmoid layer is added to normalize the weight.
The network architecture can be described as,
$$G(x)=\text{softmax}(\text{top\_k}(H(x)))$$
$H(x)$ represents the mapping function from the input to patch weight. $G(x)$ is the patch selection function.
Different architectures for the tiny network are tried and we find a simple linear layer would work decently most of the time.
Best accuracy is achieved when $k$ = 105. 
Interestingly when $k$=1, our dynamic patch selection performs worse than the fixed patch selection, implying the gated network's inability to learn useful information for decision making.

\noindent\textbf{Sequential patch weighting.} Recurrent neural networks (RNN) is specialized at processing sequential data like text and speech.
 RNN has been widely adopted as an attention mechanism to focus on different parts sequentially. We instead employ RNN for sequential patch weighting, aiming to discover different patches for decision making.
We employ a one-layer Long Short Term Memory (LSTM) network with 512 nodes. Each node has a hidden layer of size 1024. 
The last output of sequence is selected as the final output.
We get 82.7\% in this experiment and this confirms the effectiveness of the LSTM network.

\noindent\textbf{Static patch weighting} The final and most effective method we tried is the MLP network.
The MLP network contains one hidden layer with 1024 parameters,
followed by the batch normalization layer, ReLU layer, and then the output layer.
On CUB our final accuracy is 88.7\% , 1.2\% higher than the current state-of-the-art result.
We combine pairs patch with single keypoint patch and achieves a new state-of-the-art 89.2\% accuracy.
We compare our result with several other strong baselines in Table-~\ref{class-table}.

We test our algorithm also on the NAbirds dataset and the result is shown in~\ref{nabirds_table}. Our algorithm attains an accuracy of 87.9\%, more than 8\% better than a strong baseline. 

\begin{table}
\centering
\caption{ \textbf{Classification score on CUB.} 
 Annotation key as follows: GT = class labels; BB = bounding box annotation; KP = keypoint annotations; WEB = images downloaded from the Internet.}  
\label{class-table}
\begin{tabular}{|l | c | c |}

\hline
& Annotations & Accuracy \\
\hline
\hline
Huang \etal~\cite{HuangXTZ_CVPR2016} & GT+BB+KP & 76.2 \\
Zhang \etal~\cite{ZhangDGD_ECCV2014} & GT + BB & 76.4 \\
Krause \etal~\cite{KrauseJYF_CVPR2015} & GT+BB & 82.8 \\
Jaderberg \etal~\cite{JaderbergSZK_NIPS2015}  & GT & 84.1 \\
Shu \etal~\cite{KongF_CVPR2017} & GT & 84.2 \\
Zhang \etal~\cite{ZhangXZLT_CVPR2016} & GT & 84.5 \\
Xu \etal~\cite{XuHZT_ICCV2015} & GT+BB+KP+WEB & 84.6 \\
Lin \etal~\cite{LinRM_ICCV2015} & GT+BB & 85.1 \\
Cui \etal~\cite{CuiZWLLB_CVPR2017} & GT & 86.2 \\
Lam \etal~\cite{LamMT_CVPR2017}& GT+KP & 87.5 \\
\hline
\hline
PAIRS Only & GT + KP & 88.7 \\
PAIRS+Single & GT + KP & \textbf{89.2} \\
\hline

\end{tabular}

\end{table}

\begin{table}[h]
    \begin{center}
     \caption{Performance Comparison on the NABirds dataset.}
\begin{tabular}{|l|c|}
	\hline
	Algorithm & Accuracy\\
	\hline
    ResNet-50 Baseline & 79.2\% \\
	\hline
    Bilinear CNN (PAMI 2017)~\cite{LinRM_PAMI2017} & 79.4\% \\
    \hline
    {\bf PAIRS} & \bf{87.9\%} \\
	\hline
\end{tabular}

\label{nabirds_table}
\end{center}
\end{table}

\begin{figure}[h]
        \centering
\includegraphics[width=0.8\linewidth]{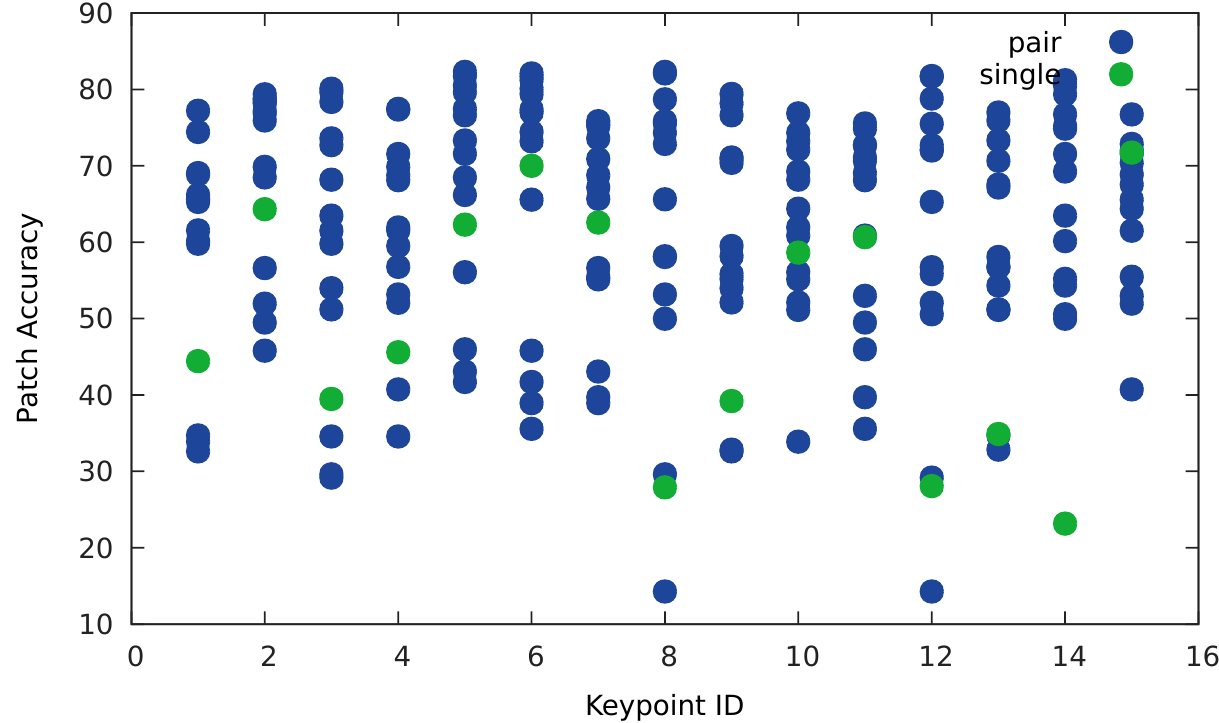}
        
        \caption{ {\textbf{Patch Accuracy: Axis-Aligned Patches v.s. PAIRS Patches}}
        We compare the patch accuracy of single keypoint patches of keypoint \emph{k} and PAIRS patches containing \emph{k}. The x axis is the keypoint id and the y axis is the patch accuracy. Most single keypoint patch are inferior to PAIRS patches in terms of patch accuracy.}
        \label{fig:scatter}
\end{figure}

\subsection{Ablation Study}

\textbf{Axis-Aligned v.s. Pose-Aligned}
We compare and show patch classification accuracy using pose-aligned patches v.s. single keypoint based axis-aligned patches in Figure~\ref{fig:scatter}. Single keypoint based patches performs consistently poorly compared to the pairs patches, confirming the effectiveness of disentangled feature representation.

\textbf{Patch Size Study}
One hyper-parameter in our algorithm is the pose-aligned patch size. We tries several size options on the best performing patch. We see that the larger-size patches generally yield better accuracy. We adopt $256 \times 512$ empirically because our base model is pretrained for such size.

\textbf{Choice of Pose Estimation Network} To test the influence of pose estimation network on the proposed algorithm, we train a separate Stacked Hourglass Network~\cite{newell2016stacked} for comparison. Stacked Hourglass Network is about ~2\% better than the FCN on PCK score, but the final classification accuracy numbers are comparable.

\subsection{Results Visualization}

We show the patch classification accuracy for each patch in the CUB dataset in Figure~\ref{fig:many_graph}. The best performing patch corresponds to \patchpair{belly}{crown}, achieving 79.6\% accuracy. The worst performing patch is the \patchpair{left-leg}{right-leg} pair which achieves only  15.7\% accuracy. Empirically, global patches perform better than local patches, however local patches are also important for localizing discriminative object parts. Patches found by beam search, as shown in Figure~\ref{fig:many_graph}, can provide insight -- a combination of global and local patches are selected to achieve an optimal result.

As hard cases often can only be classified by a few highly localized discriminative parts, the number of patches with correct predictions reflects the difficulty of the image. We propose to use the correctly predicted patch number as the indicator of the image difficulty. This is a histogram reflecting the count of many images have a given number of patches correctly predicted the class (top right plot in Figure~\ref{fig:many_graph}). Example images, ranging from hard on the left, to easy on the right, are shown below; the hard cases can be due either to very similar/easily confused classes or to pose-estimation failure.

\section{Conclusion}

Fine-grained recognition is an area where computer algorithms can assist humans on difficult tasks like recognizing bird species. Pose variation constitutes a major challenge in fine-grained recognition that recent works fail to address. In this paper, we introduce a unified object representation built from pose-aligned patches which disentangle the appearance features from the influence of pose, scale and rotation. Our proposed algorithm attains state-of-the-art performance on two fine-grained datasets, suggesting the critical importance of disentangling pose and appearance in fine-grained recognition.


\balance

{\small
\bibliographystyle{ieee}
\bibliography{FarrellMendeley,addl_refs}
}

\end{document}